# MISeval: a Metric Library for Medical Image Segmentation Evaluation

Dominik MÜLLER[a,b,1], Dennis HARTMANN[a], Philip MEYER[a,b],
Florian AUER[a], Iñaki SOTO-REY[b] and Frank KRAMER[a]
[a] *IT-Infrastructure for Translational Medical Research, University of Augsburg, Germany*
[b] *Medical Data Integration Center, Institute for Digital Medicine, University Hospital Augsburg, Germany*

**Abstract.** Correct performance assessment is crucial for evaluating modern artificial intelligence algorithms in medicine like deep-learning based medical image segmentation models. However, there is no universal metric library in Python for standardized and reproducible evaluation. Thus, we propose our open-source publicly available Python package MISeval: a metric library for Medical Image Segmentation Evaluation. The implemented metrics can be intuitively used and easily integrated into any performance assessment pipeline. The package utilizes modern CI/CD strategies to ensure functionality and stability. MISeval is available from PyPI (miseval) and GitHub: https://github.com/frankkramer-lab/miseval.

## 1. Introduction

In the last decade, computer vision analysis based on artificial intelligence methods like deep learning has seen rapid growth in prediction capabilities [1]. This resulted in clinicians striving to integrate computer vision algorithms, like image segmentation, into the medical field. Medical image segmentation (MIS) covers the automated identification and annotation of medically relevant regions of interest (ROI), which can be organs, cell structures, or medical abnormalities like tumors [2]. The idea, especially in radiology and pathology, is to establish these MIS methods in their clinical routine to reduce time-consuming processes and to aid in diagnosis as well as treatment decisions [1]. However, due to the direct impact on medical decisions, the correct evaluation of MIS models is crucial. Nevertheless, recent studies indicated widespread statistical bias in evaluations of MIS models which is also caused by incorrect metric implementation [3]. Furthermore, to our knowledge, there is no universal metric library in Python for standardized and reproducible evaluation.

In this work, we propose our open-source publicly available Python package MISeval, which is a metrics library for correct MIS model evaluation. It facilitates an intuitive and fast usage of various popular metrics from literature, as well as ensures implementation functionality and stability.

## 2. Methods

The open-source Python module MISeval is a metric library for **M**edical **I**mage **S**egmentation **Eval**uation. The library contains various commonly used metrics for image segmentation, which can be easily imported and instantly used for model performance assessment. MISeval is structured as an API with a central core interface for intuitive usage and is implemented in the programming language Python, which is platform-independent and highly popular for computer vision tasks. This allows simple and fast integration of MISeval in commonly used platforms like Tensorflow, PyTorch, or any NumPy-compatible image segmentation pipeline.

### 2.1. Metric library

Over the last decades, the MIS literature introduced a large variety of metrics for evaluation. Especially for semantic segmentation, model performance assessment can be quite complex due to the need for scoring pixel classification as well as localization correctness between predicted and annotated segmentation.

A summary, of all implemented metrics in MISeval, can be seen in Table 1. The defined formulas are based on the computation of a confusion matrix for a binary segmentation task, which contains the number of true positive (TP), false positive (FP), true negative (TN), and false negative (FN) predictions. Exceptions to this are the Adjusted Rand Index, where $n_{i,j}$, $a_i$ and $b_j$ are defined as number of pairs of elements in the confusion matrix, and the Hausdorff distance, where $h(A, B)$ as well as $d(A, B)$ is the directed Hausdorff as well as directed average Hausdorff distance, respectively [4]. In the Hausdorff distance, A and B represent the ground truth as well as predicted segmentation, respectively, and $\|a-b\|$ represents a distance function like Euclidean [4].

### 2.2. Core Interface: Evaluate()

The core of our package is the *evaluate()* function, which acts as a simple and intuitive interface to access and run all implemented metrics. The desired backbone metric for the *evaluate()* function can be defined by passing the name of an already implemented metric or by passing a user-created metric function for uncomplicated integration of custom metrics. Moreover, our core function handles automatically binary as well as multi-class problems. This allows straightforward passing of any ground truth and predicted segmentation masks to the *evaluate()* function for computing the metric assessment in a single line of code.

### 2.3. Package Stability: CI/CD

Our MISeval package utilizes modern DevOps strategies to ensure package stability and functionality during ongoing development [5]. After each update, the source code is automatically built in a reproducible environment (continuous integration – CI), extensively tested via unit testing, released, and, finally, deployed in the scientific community's MIS projects (continuous deployment – CD).

**Table 1.** Summary of currently implemented and usable metrics in MISeval. For in-detail formula description and theory of the presented metrics, we refer to the excellent review from Taha et al. [4].

| Metric | Formula | |
|---|---|---|
| Dice Similarity Coefficient / F1-score | $DSC = \dfrac{2TP}{2TP + FP + FN}$ | (1) |
| Intersection-Over-Union / Jaccard Index | $IoU = \dfrac{TP}{TP + FP + FN}$ | (2) |
| Sensitivity | $Sens = \dfrac{TP}{TP + FN}$ | (3) |
| Specificity | $Spec = \dfrac{TN}{TN + FP}$ | (4) |
| Precision | $Precision = \dfrac{TP}{TP + FP}$ | (5) |
| Accuracy / Rand Index | $Acc = \dfrac{TP + TN}{TP + TN + FN + FP}$ | (6) |
| Balanced Accuracy | $BACC = \dfrac{Sensitivity + Specificity}{2}$ | (7) |
| Adjusted Rand Index | $ARI = \dfrac{\sum_{i,j}\binom{n_{i,j}}{2} - [\sum_i \binom{a_i}{2} \sum_j \binom{b_j}{2}]/\binom{n}{2}}{\frac{1}{2}[\sum_i \binom{a_i}{2} + \sum_j \binom{b_j}{2}] - [\sum_i \binom{a_i}{2} \sum_j \binom{b_j}{2}]/\binom{n}{2}}$ | (8) |
| AUC | $AUC = 1 - \dfrac{1}{2}\left(\dfrac{FP}{FP + TN} + \dfrac{FN}{FN + TP}\right)$ | (9) |
| Cohen's Kappa | $f_c = \dfrac{(TN + FN)(TN + FP) + (FP + TP)(FN + TP)}{TP + TN + FN + FP}$ $Kap = \dfrac{(TP + TN) - f_c}{(TP + TN + FN + FP) - f_c}$ | (10) |
| Hausdorff Distance | $h(A, B) = \max_{a \in A} \min_{b \in B} \|a - b\|$ $HD(A, B) = \max(h(A, B), h(B, A))$ | (11) |
| Average Hausdorff Distance | $d(A, B) = \dfrac{1}{N} \sum_{a \in A} \min_{b \in B} \|a - b\|$ $AHD(A, B) = \max(d(A, B), d(B, A))$ | (12) |
| Volumetric Similarity | $VS = 1 - \dfrac{|FN - FP|}{2TP + FP + FN}$ | (13) |

Our unit testing considers functionality, edge cases, and exceptions for each metric. For application (functionality and edge cases), multiple dummy dataset types like empty, full, or random segmentation masks as well as single and multi-class masks are tested in all combinations. For exception handling, cases with incorrect parameter usage and non-matching mask shapes are tested.

## 2.4. Package Availability

The MISeval package is hosted, supported, and version-controlled in the Git repository platform GitHub. This allows the utilization of platform-hosted CI/CD workflows and a hub for package documentation, community contributions, bug reporting as well as feature requests. The Git repository is available under the following link:

*https://github.com/frankkramer-lab/miseval*. Furthermore, MISeval is published in the Python Package Index (PyPI), which is the official third-party software repository for Python. Thus, MISeval can be directly installed and immediately used in any Python environment using "*pip install miseval*".

Our code is licensed under the open-source GNU General Public License Version 3 (GPL-3.0 License), which allows free usage and modification for anyone.

## 3. Results

For functionality demonstration, we setup a deep-learning based MIS pipeline, trained a COVID-19 segmentation model for CT scans, computed predictions, and evaluated model performance using MISeval. The evaluation results are illustrated in Figure 1.

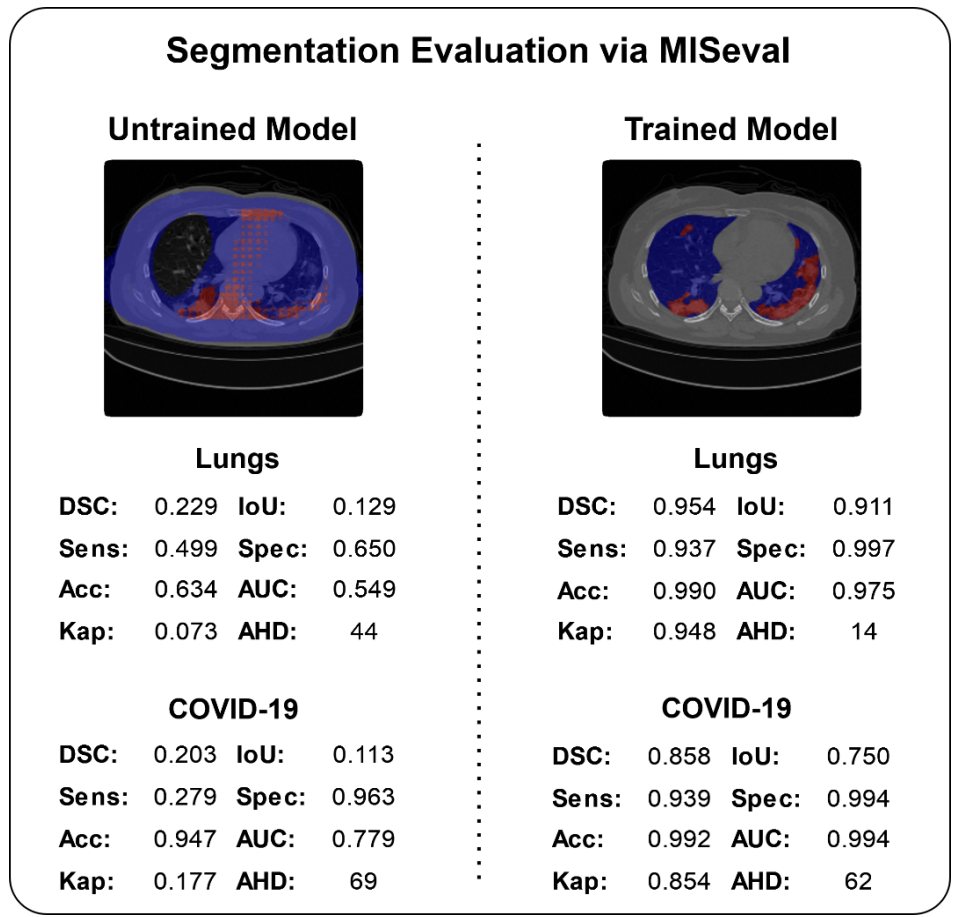

**Figure 1.** Illustration of various selected metrics from the library of MISeval to evaluate model performance for medical image segmentation on the use case COVID-19 infected region segmentation. The figure compares an untrained model (after 1 epoch during training) to a fully trained model (after 163 epochs) and shows computed tomography scans for each model with predicted lungs (blue) and infected regions (red).

The analysis utilized the MIS framework MIScnn [2] and was performed with the following parameters: Sampling in 80% training, 20% testing sets; resizing into 512x512 images; value intensity normalization via Z-score; online image-augmentation, standard U-Net architecture with focal Tversky loss function and a batch size of 24; advanced training features like dynamic learning rate, early stopping and model checkpoints. The training was performed for 163 epochs until early stopping and on 75 randomly selected images per epoch. As dataset, we used annotated computed tomography scans of COVID-19 positive patients from Ma et al. [6].

## 4. Discussion

Our proposed package MISeval allows a universal, reproducible, and standardized application of various metrics for MIS evaluation, which hopefully reduces the risk of statistical bias in studies through incorrect custom implementations. By following the state-of-the-art package stability and availability strategies, MISeval has the potential to be integrated into any future scientific performance analysis due to package stability, easy accessibility, and further contribution possibilities.

Our road map and future direction for MISeval is to ensure ongoing support, the further extension of our metric library, and providing guidelines on correct metric usage as well as evaluation. Furthermore, we plan to propose a new metric similar to the Dice Similarity Coefficient for handling the current issue of evaluating non-present classes in ground truth annotations like in control samples.

## 5. Conclusions

In this work, we proposed our open-source Python package MISeval: a metric library for medical image segmentation evaluation. The library contains various popular metrics which can be easily used and integrated into any performance assessment for image segmentation models. MISeval can be directly installed as a Python library from PyPI (miseval) and is available in GitHub: *https://github.com/frankkramer-lab/miseval*.